\documentclass{article} 
\usepackage{iclr2026_conference,times}

\usepackage{amsmath,amsfonts,bm}

\def\eqref#1{equation~\ref{#1}}

\def\1{\bm{1}}

\def\va{{\bm{a}}}

\def\vx{{\bm{x}}}

\def\mA{{\bm{A}}}
\def\mB{{\bm{B}}}

\def\mG{{\bm{G}}}

\def\mI{{\bm{I}}}

\def\mP{{\bm{P}}}

\def\mW{{\bm{W}}}

\DeclareMathAlphabet{\mathsfit}{\encodingdefault}{\sfdefault}{m}{sl}
\SetMathAlphabet{\mathsfit}{bold}{\encodingdefault}{\sfdefault}{bx}{n}

\newcommand{\E}{\mathbb{E}}

\newcommand{\R}{\mathbb{R}}

\usepackage[pagebackref=true,breaklinks=true,colorlinks=true,bookmarks=false]{hyperref}

\definecolor{deepred}{HTML}{940000}
\hypersetup{linkcolor=deepred}
\hypersetup{urlcolor  = [rgb]{0.4,0.15,0.95}}
\hypersetup{citecolor=[rgb]{0.4,0.15,0.95}}
\usepackage{url}

\usepackage{enumitem}
\usepackage{graphicx}
\usepackage{amsmath}
\usepackage{caption}
\usepackage{subcaption}
\usepackage{graphicx} 
\usepackage{multirow}
\usepackage{bm}

\usepackage{listings}
\lstdefinestyle{Pytorch}{
    language = Python,
    backgroundcolor = \color{white},
    basicstyle = \fontsize{9pt}{8pt}\selectfont\ttfamily\bfseries,
    columns = fullflexible,
    aboveskip=1pt,
    belowskip=1pt,
    breaklines = true,
    captionpos = b,
    commentstyle = \color{codeblue},
    keywordstyle = \color{codekw},
}
\usepackage{anyfontsize}
\usepackage{booktabs}
\usepackage{wrapfig}
\usepackage{tikz}
\usepackage{pgfplots}
\usepackage[dvipsnames]{xcolor}
\definecolor{brickred}{RGB}{203,65,84}
\definecolor{darkcyan}{RGB}{0,139,139}
\definecolor{Gray}{gray}{0.92}
\usepackage{fontawesome5}

\usepackage{algorithm}
\usepackage{amssymb}
\usepackage{tcolorbox}
\definecolor{metablue}{RGB}{24,119,242}
\usepackage[table]{xcolor}
\usepackage{xspace}

\usepackage{adjustbox}
\usepackage{threeparttable}
\usepackage{array}      
\usepackage{siunitx}
\usepackage{tabularx}
\newcommand{\pos}[1]{$_{\textcolor{green!60!black}{+#1}}$}
\newcommand{\nega}[1]{$_{\textcolor{red!75!black}{-#1}}$}

\newcommand{\ie}{\emph{i.e.}}

\usepackage{caption}

\usepackage[align=center,shadow=true,shadowsize=4.5pt,nobreak=true,framemethod=tikz,skipabove=9.5pt,skipbelow=9pt,innertopmargin=5pt,innerbottommargin=5pt,innerleftmargin=5pt,innerrightmargin=5pt,leftmargin=1.5pt,rightmargin=1.5pt]{mdframed}
\usetikzlibrary{shadows}

\newlength\savewidth\newcommand\shline{\noalign{\global\savewidth\arrayrulewidth
  \global\arrayrulewidth 1pt}\hline\noalign{\global\arrayrulewidth\savewidth}}

\title{Normalized Low-Rank Adaptation}

\author{
Jiale Kang$^{1,3}$ \quad
Ziyin Yue$^{}$ \quad
Zheng Zhan$^{2}$ \quad
Yangyi Huang$^{3}$\quad
Weiyang Liu$^{3,4}$ \\[2pt]
$^{1}$Yuanshi Intelligence \quad $^{2}$Microsoft Research \quad $^{3}$The Chinese University of Hong Kong \quad \\$^{4}$Shenzhen Loop Area Institute
}

\iclrfinalcopy 
\begin{document}

\maketitle

{
\vspace{-11mm}
\begin{center}
        \fontsize{9pt}{\baselineskip}\selectfont
        {\tt\href{https://spherelab.ai/NoRA}{\textbf{spherelab.ai/NoRA}}}
        \vspace{2mm}
\end{center}
}

\begin{abstract}


While low-rank adaptation (LoRA) is widely used for parameter-efficient model adaptation, how to regularize its training dynamics for stable and effective optimization remains underexplored. Because LoRA initializes the up-projection to zero, its early optimization dynamics are largely governed by the down-projection. Building on this observation, we introduce Normalized Low-Rank Adaptation (NoRA), a simple yet effective method that normalizes the down-projection matrices during training. We further show that the same normalization can be applied only at initialization, improving standard LoRA without requiring repeated normalization throughout training. Across pretraining, supervised finetuning, and reinforcement learning, NoRA consistently accelerates convergence, improves performance and training stability, and mitigates catastrophic forgetting. These benefits require neither additional trainable parameters nor inference-time computation, making NoRA a simple and broadly applicable enhancement to LoRA.

\end{abstract}

\section{Introduction}

As large language models (LLMs) continue to scale, their training costs grow rapidly, increasing the need for parameter-efficient learning. Among parameter-efficient finetuning (PEFT) approaches, Low-rank adaptation (LoRA)~\citep{hu2021lora} has become particularly popular due to its simplicity, efficiency, and strong empirical performance. By representing weight updates through low-rank factorization, LoRA greatly reduces the number of trainable parameters and has been successfully applied to supervised finetuning, pretraining, and reinforcement learning~\citep{wang2025tinatinyreasoningmodels,shen2026geometry}.
Despite its effectiveness, LoRA’s optimization dynamics remain poorly understood. Standard LoRA parameterizes
$\Delta\bm{W}=\alpha\bm{B}\bm{A}$ with randomly initialized $\bm{A}$ and zero-initialized $\bm{B}$. Although this preserves the pretrained initialization, early optimization depends entirely on the random features induced by $\bm{A}$.
This observation motivates the question below:
\vspace{3mm}
\begin{mdframed}[
    backgroundcolor=gray!4,
    linecolor=gray!35,
    linewidth=0.5pt,
    topline=true,
    bottomline=true,
    rightline=true,
    leftline=true,
    innerleftmargin=8pt,
    innerrightmargin=8pt,
    innertopmargin=6pt,
    innerbottommargin=5pt,
    skipabove=0pt,
    skipbelow=0pt,
    roundcorner=2pt
]
\centering
\fontsize{9.3pt}{11.2pt}\selectfont
\itshape
\textcolor{black!82}{
Does the regularization of the down-projection matrix $\bm{A}$ provide a useful design dimension\\for improving LoRA optimization?
}
\end{mdframed}
\vspace{-2mm}

Existing work suggests that initialization can substantially affect low-rank optimization. Methods such as PiSSA~\citep{meng2024pissa} and MiLoRA~\citep{wang2025milora} construct the initial low-rank subspace from the spectral structure of pretrained weights and often converge faster than random initialization. However, they require singular-value decomposition and modify the initial decomposition of the pretrained weights, making them less convenient for large-scale training and potentially fragile in reinforcement-learning settings~\citep{yin2025evaluatingparameterefficientmethods}. Other approaches improve LoRA through modified scaling~\citep{kalajdzievski2023rank}, constrained parameterizations~\citep{liu2024dora}, or fixed projections~\citep{kopiczko2024vera}. However, how the regularization of the down-projection matrix affects or improves LoRA's optimization remains poorly understood.

Our investigation begins with the observation that the normalized latent bottleneck used in Multi-head Latent Attention (MLA) substantially stabilizes training and also improves convergence. Similar observations have also been made in multiple training settings~\citep{liu2017deep,liu2018decoupled,loshchilov2025ngpt}. These phenomena suggest that the numerical structure of the normalized latent representation presented to the up-projection can play an important role in stable optimization. Directly introducing latent normalization into LoRA changes the standard linear update from $\bm B\bm A\bm x$ to $\bm{B} \cdot \text{Norm}(\bm A\bm x)$. Because the normalization depends on the input, the resulting transformation is nonlinear with respect to $\bm x$ and can no longer be exactly merged into the pretrained weight matrix like LoRA. Inspired by MLA, we therefore study whether the useful numerical effect of latent normalization can instead be encoded as a regularization  of the linear down-projection such that we can have both LoRA's exact mergeability and MLA's favorable optimization benefits.

To this end, we adapt MLA's normalization principle from the latent feature to the low-rank projection itself. Each column of $\bm{A}\in\mathbb R^{r\times k}$ represents how an input coordinate is projected into the $r$-dimensional latent space. Guided by the normalization dimension in MLA, we propose to normalize these projection vectors along the rank dimension, yielding the linear form $\bm{B}\cdot\text{Norm}(\bm{A})\bm{x}$. We refer to this formulation as Normalized LoRA (NoRA), which constrains the input-to-latent projection magnitudes throughout training while preserving exact weight mergeability. 
This normalization scheme also ensures that the gradient norm of NoRA can be better aligned with full finetuning, bridging the learnability gap between LoRA and full finetuning.


NoRA requires the down-projection matrix $\bm{A}$ to be normalized along the output rank dimension. This principle motivates two ways of applying normalization: (1) as a training constraint (\ie, NoRA) and (2) as an initialization strategy (\ie, NoRA-init). In the former, NoRA continuously normalizes the down-projection matrix $\bm{A}$ along the low-rank dimension $r$ throughout training. In the latter, NoRA-init normalizes $\bm{A}$ only once at initialization and subsequently optimizes it without constraints during training. For NoRA-init, we propose two initialization variants: (1) random normalization, where $\bm{A}$ is randomly initialized and then normalized; and (2) Block Identity Matrix Initialization (BIMI), where $\bm{A}$ is constructed as a block-identity matrix that inherently satisfies the rank-dimensional normalization requirement. NoRA improves training by removing undesirable scale imbalance across the low-rank dimensions, providing a better-conditioned parameterization than standard LoRA.

We evaluate both NoRA and NoRA-init across pretraining, supervised finetuning, and reinforcement learning with verifiable rewards. Across a diverse range of base models, tasks, and training configurations, both methods consistently accelerate convergence, improve training stability, and enhance downstream performance. They also mitigate catastrophic forgetting during supervised adaptation and remain robust in reinforcement-learning settings where other spectral initialization methods can be fragile. In general, we can conclude that these results establish rank-dimension normalization of the down-projection as an important design principle for improving LoRA optimization.
Our contributions are summarized as follows:

\vspace{0.5mm}
\begin{itemize}[leftmargin=4mm,nosep]
\setlength\itemsep{0.4em}
\item We identify the down-projection matrix $\bm{A}$ as a previously underexplored yet important design dimension in LoRA, showing that its magnitude can play a critical role in shaping optimization dynamics and ultimately improving training efficiency and downstream performance.
\item We propose NoRA, which normalizes the down-projection matrix $\bm{A}$ along the rank dimension in the LoRA parameterization. We further introduce NoRA-init, an initialization-only variant that applies this normalization at initialization and requires no re-normalization during training, enabling seamless performance improvements without modifying the optimization procedure.
\item Extensive experiments across pretraining, SFT, and RLVR demonstrate NoRA's improved convergence, stability, downstream performance, and resistance to catastrophic forgetting. 
\end{itemize}

\section{Preliminaries}

LoRA parameterizes the weight update of a pretrained matrix as
\begin{equation}
\Delta\bm W=\bm{W}-\bm{W}_0=\alpha\bm B\bm A,
\end{equation}
where
$\bm A\in\mathbb R^{r\times k}$,
$\bm B\in\mathbb R^{d\times r}$,
and
$r\ll\min(d,k)$.
Standard LoRA initializes
$\bm B^{(0)}=\bm0$
to preserve the pretrained model.
Let
$\bm G=\partial\mathcal L/\partial\bm W$
denote the gradient of the corresponding full weight matrix.
At initialization, we have the following gradients with respect to $\bm{B}$ and $\bm{A}$:
\begin{equation}
\left.
\frac{\partial\mathcal L}{\partial\bm B}
\right|_{t=0}
=
\alpha
\bm G
(\bm A^{(0)})^\top,
\qquad
\left.
\frac{\partial\mathcal L}{\partial\bm A}
\right|_{t=0}
=
\bm0.
\label{eq:init_grad}
\end{equation}
Therefore, the earliest optimization of LoRA is entirely determined by the latent representation induced by the initial projection
$\bm A^{(0)}$.

Motivated by this observation, we consider a unified latent-feature formulation:
\begin{equation}
\Delta\bm y
=
\alpha
\bm B
\bm\phi(\bm x),
\end{equation}
where
$\bm\phi(\bm x)\in\mathbb R^r$
denotes the latent feature presented to the up-projection.
Existing low-rank methods can be interpreted as constructing this latent feature in different ways:
\begin{equation}
\bm\phi(\bm x)=
\begin{cases}
\bm A\bm x,
&
\text{random projection (LoRA)},\\[2mm]
\bm A_{\rm init}\bm x,
&
\text{structured projection (e.g., PiSSA, MiLoRA)},\\[2mm]
\text{Norm}(\bm A\bm x),
&
\text{normalized projection (e.g., MLA)}.
\end{cases}
\label{eq:latent_view}
\end{equation}
Although these approaches adopt different implementations, they can all be viewed as designing the latent feature presented to the up-projection. This latent-projection perspective motivates us to revisit how to regularize the latent projection without losing the exact mergeability of LoRA.

\begin{table}[t]
\centering
\scriptsize
\renewcommand{\arraystretch}{1.22}
\setlength{\tabcolsep}{23pt}
\setlength{\abovecaptionskip}{5pt}
\setlength{\belowcaptionskip}{3pt}
\begin{tabular}{lll}
\textbf{Method} & \textbf{Forward Computation} & \textbf{Initialization / Prior} \\
\shline

\multicolumn{3}{l}{\textit{\textbf{Full-Parameter and Standard Low-Rank Adaptation}}} \\

 Full Finetuning
& $y = \bm{W}_0 x$
& N/A \\

 LoRA~\citep{hu2021lora}
& $y = \bm{W}_0 x + \bm{B}\bm{A}x$
& $\bm{A}\sim\mathcal{N}(0,\sigma^2),\ \bm{B}=0$ \\

\hline

\multicolumn{3}{l}{\textit{\textbf{Optimization-Strategy Variants}}} \\

 LoRA+~\citep{hayou2024lora+}
& $y = \bm{W}_0x + \bm{B}\bm{A}x$
& $\eta_{\bm{B}} = \lambda \eta_{\bm{A}}$ \\

\hline

\multicolumn{3}{l}{\textit{\textbf{Structured or Constrained Low-Rank Parameterizations}}} \\

 DoRA~\citep{liu2024dora}
& $y = \bm{m}\!\left(\frac{\bm{W}_0x+\bm{B}\bm{A}x}{\|\bm{W}_0+\bm{B}\bm{A}\|_c}\right)$
& Kaiming init,\ $\bm{B}=0$ \\

 MiSS~\citep{kang2026miss}
& $y = \bm{W}_0x + \operatorname{expand}(\bm{D})x$
& Fixed structured projection \\

 AdaLoRA~\citep{zhang2023adalora}
& $y = \bm{W}_0x + \bm{P}\bm{\Lambda}\bm{Q}x$
& $\bm{\Lambda}=0,\ \bm{P},\bm{Q}\sim\mathcal{N}(0,\sigma^2)$ \\

 LoRA-FA~\citep{zhang2023lora}
& $y = \bm{W}_0x + \bm{B}\bm{A}x$
& Frozen random $\bm{A},\ \bm{B}=0$ \\

 VeRA~\citep{ye2025vera}
& $y = \bm{W}_0x + \bm{\Lambda}_b\bm{B}\bm{\Lambda}_d\bm{A}x$
& Frozen random $\bm{A},\bm{B}$ \\

\hline

\multicolumn{3}{l}{\textit{\textbf{Initialization-Based Low-Rank Adaptation}}} \\

 PiSSA~\citep{meng2024pissa}
& $y = (\bm{W}_0-\bm{B}\bm{A})x+\bm{B}\bm{A}x$
& Top-$r$ SVD initialization \\

 MiLoRA~\citep{wang2025milora}
& $y = (\bm{W}_0-\bm{B}\bm{A})x+\bm{B}\bm{A}x$
& Minor-component SVD init \\
\rowcolor{Gray}
 NoRA-init
& $y = \bm{W}_0x + \bm{B}\bm{A}x$
& $\bm{A}=\text{Norm}(\bm{A}),\ \bm{B}=0$ \\
\rowcolor{Gray}
 NoRA
& $y = \bm{W}_0x + \bm{B}\operatorname{Norm}(\bm A)\bm x$
& $ \bm{B}=0$ \\
\end{tabular}
\caption{\footnotesize Overview of representative PEFT methods categorized by their primary mechanism, including forward parameterization and initialization strategy.}
\label{tab:overview}
\vspace{-2mm}
\end{table}

\section{\underline{NoRA}: Normalized Low-Rank Adaptation}

\subsection{From Normalized Latent Features to Normalized Low-Rank Projection} 

The normalized latent projection in MLA suggests that the representation presented to the up-projection plays an important role in training dynamics. Specifically, MLA constructs the low-dimensional latent feature as \begin{equation} \bm{\phi}(\bm{x}) = \text{Norm}(\bm{A}\bm{x}), \end{equation} where the normalization operator explicitly regulates the projected feature before it is mapped back to the output space. Directly introducing this operation into LoRA changes the standard linear update from $\bm{B}\bm{A}\bm{x}$ to $\bm{B}\text{Norm}(\bm{A}\bm{x})$. Since the normalization depends on the input, this transformation becomes nonlinear with respect to $\bm{x}$ and can no longer be exactly merged into the pretrained weight matrix. We therefore seek to transfer the scale-control effect from the latent feature to the down-projection matrix itself. Modern Transformer architectures are typically pre-normalized, such that the magnitude of the input $\bm{x}$ to each linear layer is already well controlled by LayerNorm. In this setting, variations in the column norms of $\bm{A}$ introduce additional scale imbalance in how different input coordinates are projected into the low-rank latent space. We therefore normalize each column of $\bm{A}$ along the LoRA rank dimension. For $\bm{A}\in\mathbb{R}^{r\times k}$, we write 
\begin{equation}
\bm{A} = \left[ \bm{a}_1,\, \bm{a}_2,\, \dots,\, \bm{a}_k \right], \qquad \bm{a}_j\in\mathbb{R}^{r}, 
\end{equation} 
where $\bm{a}_j$ denotes the latent projection vector associated with the $j$-th input coordinate. We define 
\begin{equation}
\text{Norm}(\bm{A}) = \left[ \frac{\bm{a}_1} {\max(\lVert\bm{a}_1\rVert_2,\epsilon)}, \, \dots, \, \frac{\bm{a}_k} {\max(\lVert\bm{a}_k\rVert_2,\epsilon)} \right], \label{eq:rank_dimension_normalization} 
\end{equation} 
where $\epsilon$ is a small constant for numerical stability. Thus, we have that 
\begin{equation} \left\| \left[ \text{Norm}(\bm{A}) \right]_{:,j} \right\|_2 = 1, \qquad j=1,\dots,k. \label{eq:unit_input_projection}
\end{equation} 
Based on this rank-dimension normalization, we introduce normalized low-rank adaptation, which parameterizes the low-rank update as the following form:
\begin{equation} 
\Delta\bm{y} = \alpha \bm{B}\text{Norm}(\bm{A})\bm{x}. \label{eq:nora}
\end{equation}
Unlike latent-feature normalization in MLA, $\text{Norm}(\cdot)$ depends only on the LoRA parameters rather than the input. NoRA therefore imposes a normalization constraint on the down-projection throughout training while remaining linear with respect to $\bm{x}$. After training, the normalized down-projection can be absorbed into the low-rank update, preserving exact weight mergeability. In this way, every input coordinate is continuously projected through a unit-norm latent direction, removing arbitrary column-wise scale variation in $\bm{A}$. The resulting transition can be summarized as 
\begin{equation}
\underbrace{ \bm{B}\text{Norm}(\bm{A}\bm{x}) }_{\text{\textbf{MLA}: input-dependent latent normalization}} \quad\rightarrow\quad \underbrace{ \bm{B}\text{Norm}(\bm{A})\bm{x} }_{\text{\textbf{NoRA}: normalized low-rank projection}}.
\end{equation} 
Our initialization and gradient-norm analyses further suggest that much of the benefit of NoRA is already established at the beginning of optimization. This motivates a simpler initialization-only variant, which we denote as NoRA-init. Specifically, we initialize
\begin{equation} \bm{A}^{(0)} = \text{Norm}(\bm{A}_{\text{init}}), \qquad \bm{B}^{(0)} = \bm{0}, \label{eq:nora_init} 
\end{equation} 
and subsequently optimize $\bm{A}$ and $\bm{B}$ using the standard LoRA parameterization without performing normalization throughout training. $\bm{A}_\text{init}$ denotes the random initialization in LoRA. Despite removing the persistent constraint during training, NoRA-Init retains most of the optimization benefit of NoRA in our experiments. This observation suggests that controlling the scale of the input-to-latent projection at initialization is particularly important for the early optimization of low-rank adaptation.

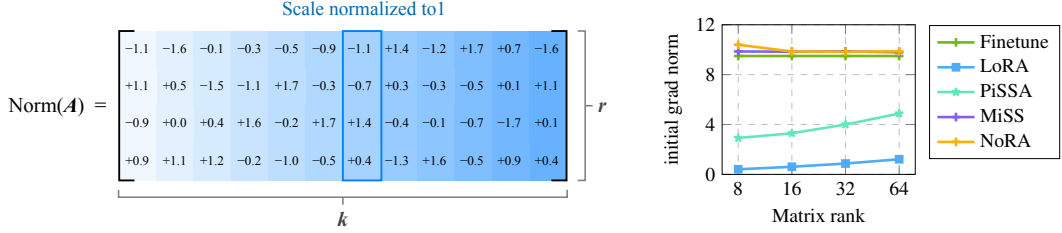
\begin{figure}[t]
\vspace{-13pt}
\centering
\hspace{-0.15\textwidth}
\begin{minipage}[t]{0.6\textwidth}
\centering
\includegraphics[width=.95\linewidth]{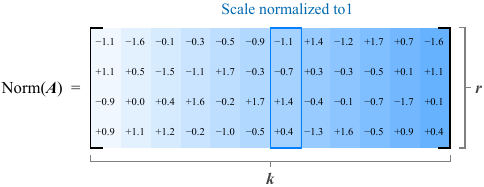}
\end{minipage}
\begin{minipage}[t]{0.275\textwidth}
\centering
\resizebox{\linewidth}{!}{\definecolor{color1}{RGB}{116, 184, 22}
\definecolor{color2}{RGB}{77, 171, 247}
\definecolor{color3}{RGB}{99, 230, 190}
\definecolor{color4}{RGB}{132, 94, 247}
\definecolor{color5}{RGB}{250, 176, 5}

    \begin{tikzpicture} 
    \small
    \begin{axis}[
        name=plot2,
        sharp plot, 
        title style={align=right},
        xlabel=Matrix rank, 
        width=4.7cm, height=4cm,  
        ymin=0, ymax=12,  
        symbolic x coords={8,16,32,64},
        ytick={0,4,8,12}, 
        yticklabels={0,4,8,12},
        xlabel near ticks, 
        ylabel near ticks, 
        ylabel=initial grad norm,
        ymajorgrids=true, 
        xmajorgrids=true, 
        grid style=dashed, 
        legend style={at={(1.75,0.55)},anchor=east, legend cell align=left, font=\small}, 
        legend style={overlay},
        clip=false,
    ]

    \addplot[very thick,mark=+,mark options={scale=1}, color=color1] plot coordinates { 
                        (8,9.494)
                        (16,9.494)
                        (32,9.494)
                        (64,9.494)
    };
    \addlegendentry{Finetune} 
    
    \addplot[very thick, mark=square*, mark options={scale=0.7}, color=color2] plot coordinates {
                        (8,0.4085)
                        (16,0.6095)
                        (32,0.8706)
                        (64,1.214)
    };
    \addlegendentry{LoRA}

    \addplot[very thick, mark=star, mark options={scale=1}, color=color3] plot coordinates {
                        (8,2.920)
                        (16,3.298)
                        (32,4.001)
                        (64,4.877)
    };
    \addlegendentry{PiSSA}  

    \addplot[very thick, mark=+, mark options={scale=1}, color=color4] plot coordinates {
                        (8,9.858)
                        (16,9.836)
                        (32,9.864)
                        (64,9.793)
    };
    \addlegendentry{MiSS}  
    \addplot[very thick, mark=+, mark options={scale=1}, color=color5] plot coordinates {
                        (8,10.40)
                        (16,9.853)
                        (32,9.831)
                        (64,9.864)
    };
    \addlegendentry{NoRA}  
    \end{axis}

    \end{tikzpicture}}
\end{minipage}
\vspace{-2mm}
\caption{\footnotesize
Illustration and optimization behavior NoRA.
\textbf{Left:} Each input-to-latent projection vector is normalized along the
LoRA rank dimension.
\textbf{Right:} Gradient-norm dynamics of different initialization methods and
LoRA ranks on LLaMA-3.2-3B trained on the Math dataset.
}
\label{fig:grad_norm}
\vspace{-.5mm}
\end{figure}

\subsection{Connection to MiSS and Block Identity Matrix Initialization}

Having established the effectiveness of rank-dimension normalization, we further investigate alternative approaches for achieving this form of normalization. MiSS~\citep{kang2026miss} is a recent PEFT method that improves parameter efficiency through a matrix-shard sharing strategy. Consider an input $\bm{x}\in\mathbb{R}^{k}$ with $k=br$, partitioned into
$b$ blocks $\bm{x}^{(i)}\in\mathbb{R}^{r}$. The MiSS update is written as
\begin{equation}
\Delta \bm{y}_{\mathrm{MiSS}}=\Delta \bm{W}\bm{x}
=
\alpha \cdot \operatorname{expand}(\bm{B})\bm{x}
=
\alpha \bm{B}\sum_{i=1}^{b}\bm{x}^{(i)}
=
\alpha \bm{B}
\underbrace{
\left[
\bm{I}_r,\,
\dots,\,
\bm{I}_r
\right]
}_{\bm{A}_{\mathrm{fix}}}
\bm{x}.
\label{eq:miss_equiv}
\end{equation}
Thus, MiSS is equivalent to a LoRA-style update with a fixed down-projection
$\bm{A}_{\mathrm{fix}}$ composed of repeated identity matrices. Importantly,
each column of this projection has unit norm:
\begin{equation}
\left\|
(\bm{A}_{\mathrm{fix}})_{:,j}
\right\|_2
=
1,
\qquad
\forall j.
\label{eq:miss_norm_property}
\end{equation}
MiSS therefore can be interpreted as a special case of NoRA, as its down-projection matrix is normalized by a block-identity construction (\ie, $\bm{A}$ is fixed to the constant $\bm{A}_{\text{fix}}$). As shown in Figure~\ref{fig:grad_norm}, NoRA produces substantially stronger early gradients than standard LoRA, with gradient norms approaching those of full finetuning. Interestingly, MiSS also exhibits a highly similar pattern. These results well justifies the empirical effectiveness of NoRA and its special case MiSS.

Motivated by the connection to MiSS, we introduce
Block Identity Matrix Initialization (BIMI) as a simple structured
realization of NoRA-init. Rather than fixing the block-identity projection throughout training, BIMI uses it only to initialize the LoRA down-projection matrix
and allows it to remain fully trainable afterward. Specifically, for
$k=br+q$, where $0\leq q<r$, we initialize
\begin{equation}
\bm{A}_{\text{bimi}}
=
\left[
\underbrace{
\bm{I}_r,\,
\bm{I}_r,\,
\dots,\,
\bm{I}_r
}_{b\ \text{blocks}},
\bm{E}_q
\right]
\in\mathbb{R}^{r\times k},
\label{eq:bimi}
\end{equation}
where $\bm{E}_q\in\mathbb{R}^{r\times q}$ contains the first $q$ columns of
$\bm{I}_r$ and is omitted when $q=0$. By construction, we have 
$\|(\bm{A}_{\text{bimi}})_{:,j}\|_2=1, \forall j,$, making BIMI directly a special case of NoRA-init. The resulting update retains the standard LoRA form, $\Delta \bm{y}=\alpha \bm{B}\bm{A}\bm{x}$,
with both $\bm{A}$ and $\bm{B}$ optimized during training.

\subsection{Why Does NoRA Work? A Preconditioning Perspective}

The early optimization of LoRA is dictated by the initialization of the down-projection matrix $\bm{A}$. This is because LoRA can be viewed as performing full-finetuning gradient descent under a hidden preconditioner that is determined by $\bm{A}$ and acts on the input coordinates. Let $\bm{G}=\partial\mathcal{L}/\partial\bm{W}\in\R^{d\times k}$ be the full-finetuning gradient. A gradient step of size $\eta$ on
$\bm{B}$ at initialization is $\Delta\bm{B}=-\eta\alpha\,\bm{G}\bm{A}^{\top}$, while
$\bm{A}$ receives no gradient and stays fixed. The induced change of the merged
weight is therefore
\begin{equation}\label{eq:precond}
\Delta\bm{W}= \alpha\,\Delta\bm{B}\,\bm{A} = -\eta\,\bm{G}\,\bm{P},
\qquad \bm{P} = \alpha^{2}\bm{A}^{\top}\bm{A}\in\R^{k\times k}.
\end{equation}
Full finetuning takes the step $\Delta\mW=-\eta\,\mG$, \ie, \eqref{eq:precond} with $\mP=\mI$. Hence, at initialization, and approximately for as long as $\mB$ remains small (the neglected term $\alpha\mB\,\Delta\mA=-\eta\alpha^{2}\mB\mB^{\top}\mG$ is second order in $\mB$), LoRA is full finetuning with the gradient right-multiplied by a positive semidefinite matrix of rank $r$ that acts on the input coordinates. Right-multiplication by a $k\times k$ matrix is exactly where curvature-based methods~\citep{martens2015optimizing} place an input-side preconditioner. Moreover, we observe that column norms are per-coordinate learning rates:
\begin{equation}\label{eq:gain}
\Delta\mW_{:,j}=-\eta\Big[\;
\underbrace{\alpha^{2}\|\va_j\|_2^{2}\;\mG_{:,j}}_{\text{coordinate $j$'s own update}}
\;+\;
\underbrace{\sum\nolimits_{i\neq j}\alpha^{2}(\va_i^{\top}\va_j)\,\mG_{:,i}}_{\text{crosstalk from the rank bottleneck}}
\Big].
\end{equation}
The first term is the full-finetuning update of the weight column with respect to input coordinate $j$, scaled by the squared length of its latent  $\bm{a}_j$. The second term is crosstalk between
input coordinates induced by the rank bottleneck, and its coefficients are inner products of latent directions.

\textbf{Random initialization of $\bm{A}$ makes learning rates unbalanced at each coordinate.}
For example, when $\mA^{(0)}$ has zero-mean i.i.d.\ entries of variance $\sigma^{2}\propto1/k$, $\E\|\va_j\|_2^{2}=r\sigma^{2}\propto r/k\ll1$. Therefore, at any moderate $\alpha$, every learning rate is far too small, so gradient flows through the adapter at a small fraction of the full-finetuning rate. This is empirically verified by the small gradient norm of LoRA in Figure~\ref{fig:grad_norm}. The learning rates are also unbalanced. $\|\va_j\|_2^{2}$ fluctuates around its mean with relative spread of order $1/\sqrt{r}$, so at small rank each coordinate gets its learning rate randomly, unrelated to data or curvature. A preconditioner should remove distortion, not add its own.

\textbf{Normalization is exactly the fix.}
Setting every $\|\va_j\|_2=1$ at unit scaling (rank-dimension normalization in NoRA), $\alpha=1$ (equivalently $\alpha=r$ under the $\alpha/r$ implementation convention), fixes the problems. Specifically, this leads to $\text{Diag}(\bm{P})=\mI$ deterministically, $\E[\bm{P}]=\bm{I}$ over the random directions, and the adapter's gradient norm can match full finetuning's gradient norm in expectation, independently of $r$. Therefore, we can get the flat NoRA curve in Figure~\ref{fig:grad_norm}. In contrast, row normalization ($\text{Norm}_k$) leaves $\text{Diag}(\bm{P})$ random of order $r/k$, and brings no empirical gain (see Table~\ref{tab:sft_init_compare}). Moreover, BIMI uses unit basis vectors as columns (the same diagonal through a completely different crosstalk pattern) and performs similarly to rank-dimension normalization of random initializations (see Table~\ref{tab:sft_init_compare}), which points to the diagonal of $\bm{P}$ as the decisive quantity.

\textbf{Initialization does most of the work; the constraint adds stability.}
Since $\mA$'s gradient, $\alpha\mB^{\top}\mG$, vanishes with $\mB$, $\mA^{(0)}$ alone fixes the learning rates and the input subspace during the decisive early phase, correcting the rates once at initialization thus captures most of the benefit. This is exactly what NoRA-init does. Performing $\text{Norm}(\mA)$ in the forward pass adds two things: (1) the loss becomes invariant to each column's scale, so gradients are tangential and, under gradient descent, norms can only grow while the effective step anneals automatically, as in weight-normalized training~\citep{liu2017deep,salimans2016weight,liu2018decoupled}; and (2) $\text{Diag}(\mP)=\mI$ is enforced for the whole run on a compact set of directions, so the latent scale can neither collapse nor explode and the update stays linear in $\vx$ and can be merged exactly. To summarize, NoRA sets the diagonal gains of LoRA's hidden preconditioner to one, and keeps them throughout training.

\vspace{-0.5mm}
\section{Experiments and Results}
\vspace{-0.5mm}
To thoroughly evaluate our method, we conduct experiments across pretraining, supervised finetuning (SFT), and reinforcement learning. We further perform ablation studies to investigate the effect of the normalization dimension and initialization distribution. Evaluation settings are in Table~\ref{tab:arena_pipeline}.

\textbf{Ablation Study.}
We investigate the effect of the normalization dimension of the down-projection matrix $\bm{A}$ by comparing column-wise and row-wise normalization, and further evaluate the effectiveness of NoRA across different random or deterministic initialization schemes.

\textbf{Pretraining.}
We begin with controlled studies on the MLA architecture under the L24-D1024 setting to examine the effect of intermediate normalization within low-rank latent representations. Building on these observations, we further evaluate NoRA on standard multi-head attention (MHA), where all attention projection layers are adapted with NoRA.

\textbf{Supervised finetuning.}
We evaluate NoRA on Llama 3.2-3B across a diverse set of downstream tasks, including mathematical reasoning and code generation, as well as its robustness against catastrophic forgetting~\citep{huang2026peft}. More detailed settings are given in Appendix~\ref{app:settings}.

\textbf{Reinforcement learning with verifiable rewards.}
To assess its effectiveness in post-training optimization, we additionally evaluate NoRA under reinforcement learning using the ReRL framework.

\begin{table}[t]
\centering
\scriptsize
\renewcommand{\arraystretch}{1.08}
\setlength{\tabcolsep}{4pt}
\renewcommand{\arraystretch}{1.24}
\setlength{\tabcolsep}{4pt}
\setlength{\abovecaptionskip}{4pt}
\setlength{\belowcaptionskip}{2pt}
\vspace{-1.5mm}
\begin{tabular}{>{\centering\arraybackslash}m{0.9cm} m{2.1cm} m{2.2cm} m{2.2cm} m{5.0cm}}
\textbf{Stage}&\multicolumn{1}{>{\centering\arraybackslash}m{2.1cm}}{\textbf{Repository}}&\multicolumn{1}{>{\centering\arraybackslash}m{2.2cm}}{\textbf{Model}}&\multicolumn{1}{>{\centering\arraybackslash}m{2.2cm}}{\textbf{Train Data}}&\multicolumn{1}{>{\centering\arraybackslash}m{5.0cm}}{\textbf{Benchmark}} \\
\shline
\textbf{Pretrain}
&
FLA~\citep{yang2024fla} 
&
MLA~\citep{deepseekai2024deepseekv2strongeconomicalefficient}\newline
MHA~\citep{vaswani2017attention} 
&
SlimPajama~\citep{cerebras2023slimpajama}
&
LAMBADA~\citep{paperno2016lambada}\newline 
WikiText~\citep{merity2016pointer} \newline 
Arc-Easy/Arc-Challenge~\citep{clark2018think}  \newline
HellaSwag~\citep{zellers2019hellaswag}\newline 
PIQA~\citep{bisk2020piqa}\newline 
OpenBookQA ~\citep{mihaylov2018can}\newline 
WinoGrande~\citep{sakaguchi2021winogrande}
\\
\hline
\textbf{SFT}
&
PEFT-Arena~\citep{huang2026peft}
&
Llama-3.2-3B~\citep{grattafiori2024llama}
&
MetaMath~\citep{yu2024metamath} \newline
CodeFeedback~\citep{zheng2024opencodeinterpreter}
&
\textbf{Math:} \newline
GSM8k ~\citep{cobbe2021training}\newline
Math500 ~\citep{yu2024metamath} \newline
\textbf{Code:}\newline
HumanEval ~\citep{chen2021evaluating} \newline
Mbpp  ~\citep{austin2021program}
\\
\hline
\textbf{RL}
&
PeRL~\citep{yin2025evaluatingparameterefficientmethods}
&
DeepSeek-R1-Distill-Qwen1.5B ~\citep{deepseekai2024deepseekv2strongeconomicalefficient}  
&
open-r1/DAPO-Math-17k-Processed~\citep{yu2025dapo}
&
AIME24~\citep{aime25} ($\mathrm{Avg}@32$) \newline
AIME25 ~\citep{aime25} ($\mathrm{Avg}@32$) \newline
MATH500~\citep{lightman2024let} ($\mathrm{Avg}@4$) \newline
Minerva~\citep{lewkowycz2022solving} ($\mathrm{Avg}@4$) \newline
AMC~\citep{li2024numinamath} ($\mathrm{Avg}@32$) \newline
HMMT~\citep{balunovic2025matharena} ($\mathrm{Avg}@32$)
\\
\end{tabular}%
\caption{
\footnotesize Overview of all the training settings in our experiments.
}
\label{tab:arena_pipeline}
\vspace{-2.5mm}
\end{table}

\vspace{-0.5mm}
\subsection{Effect of the Normalization Dimension}
\vspace{-0.5mm}

\setlength{\columnsep}{13pt}
\begin{wraptable}{r}[0cm]{0pt}
\scriptsize
\centering
\hspace{-2.4mm}
\setlength{\tabcolsep}{5pt}
\renewcommand{\arraystretch}{1.28}
\begin{tabular}{llccc}
\specialrule{0em}{0pt}{-12.5pt}
\textbf{Initialization}
& \textbf{Method}
& \textbf{GSM8K}
& \textbf{Math}
& \textbf{Avg.} \\
\shline

\multirow{3}{*}{$\mathcal U(-1/\sqrt{k},1/\sqrt{k})$}
& None
& 47.68& 10.86& 29.27 \\
& $\text{Norm}_k$
& 48.67& 10.82& 29.75 \\
& $\text{Norm}_r$
& \textbf{60.12}& 14.20& \textbf{37.16} \\
\hline
\multirow{3}{*}{$\mathcal N(0,1/r^2)$}
& None
& 50.41& 11.68& 31.05 \\
& $\text{Norm}_k$
& 49.73& 11.34& 30.54 \\
& $\text{Norm}_r$
& \underline{59.96}& 13.93& 36.95 \\
\hline
\multirow{3}{*}{$\mathcal U(-1,1)$}
& None
& 48.19& 11.02& 29.61 \\
& $\text{Norm}_k$
& 48.29& 10.86& 29.58 \\
& $\text{Norm}_r$
& 58.98& \textbf{14.34}& 36.66 \\
\hline
$\left[
\bm{I}_r,\,
\dots,\,
\bm{I}_r
\right]$
& BIMI
& 59.89& \underline{14.24}& \underline{37.07} \\
\specialrule{0em}{-4.5pt}{0pt}
\end{tabular}
\caption{\footnotesize Effect of different normalization dimensions under supervised finetuning. The average is computed over accuracies on GSM8K and Math.}
\label{tab:sft_init_compare}
\end{wraptable}

We first study whether the effect of the normalization dimension under different initialization schemes. Given $\bm A\in\mathbb R^{r\times k}$, $\text{Norm}_k$ normalizes each row, while $\text{Norm}_r$ normalizes each column $\bm A_{:,j}\in\mathbb R^r$. As shown in Table~\ref{tab:sft_init_compare}, $\text{Norm}_k$ provides little improvement, whereas $\text{Norm}_k$ consistently improves Random Uniform, Gaussian, and Kaiming Uniform initializations. After applying NoRA, all three initialization schemes achieve similar performance, indicating that its benefit is largely independent of the initialization distribution. BIMI achieves comparable performance and satisfies NoRA by construction, serving as a structured and deterministic instance of NoRA.

\vspace{-1mm}
\subsection{LLM Pretraining}
\vspace{-0.5mm}

We first conduct LLM pretraining experiments using the standard MLA architecture. Normalizing the latent representation $\bm{A}\bm{x}$ consistently improves model performance and accelerates convergence during pretraining. Motivated by these results, we extend the same normalization principle to the linear low-rank formulation by applying normalization to the down-projection $\bm{A}$, yielding NoRA. To evaluate the effectiveness of NoRA for LLM pretraining, we adopt NoRA-init and compare it with the standard low-rank parameterization. As shown in Table~\ref{tab:main_results}, NoRA-init achieves faster convergence and better downstream performance. Although its performance remains slightly below that of directly normalizing $\bm{A}\bm{x}$, NoRA-init preserves the linear structure $\bm{B}\bm{A}\bm{x}$, thereby retaining the parameter mergeability and deployment efficiency of conventional low-rank parameterizations.


We further evaluate NoRA-init under the standard MHA architecture for LLM pretraining. Specifically, we parameterize all linear projections in the attention module, including $\bm{q}_{\mathrm{proj}}$, $\bm{k}_{\mathrm{proj}}$, $\bm{v}_{\mathrm{proj}}$, and $\bm{o}_{\mathrm{proj}}$, using low-rank factors with rank $r=64$. As shown in Table~\ref{tab:main_results}, the standard low-rank parameterization exhibits a severe performance collapse on LAMBADA, along with substantial degradation on several other benchmarks. An examination of the optimization dynamics reveals that its gradient norms diminish to abnormally low levels during training, suggesting insufficient optimization. In contrast, NoRA-init achieves significantly better performance than standard low-rank parameterization and preserves well-scaled gradient magnitudes throughout training, thereby avoiding such optimization collapse and yielding  more stable loss trajectories and downstream performance.

This observation is related to the capacity-limited behavior of low-rank adaptation discussed in \cite{schulman2025lora}, where insufficient low-rank capacity can largely constrain the flexibility of optimization trajectory. Our results further show that capacity alone does not fully determine optimization behavior. Under the same rank constraint, modifying the initialization of the low-rank projection substantially changes the training dynamics. This suggests that controlling the scale and structure of the input-to-latent projection provides a more favorable optimization geometry under limited-rank parameterizations, thereby alleviating performance degradation in pretraining.

\begin{table}[t!]
\centering
\scriptsize
\renewcommand{\arraystretch}{1.26}
\setlength{\tabcolsep}{5.5pt}
\setlength{\abovecaptionskip}{5pt}
\setlength{\belowcaptionskip}{3pt}
\begin{tabular}{lccccccccccc}
\multirow{2}{*}{\textbf{Model \& Scale}}&\multirow{2}{*}{\textbf{Parameters}}
& \textbf{Lamb.$\dagger$} & \textbf{Wiki.} & \textbf{Lamb.} & \textbf{ARC$_e$} & \textbf{ARC$_c$} & \textbf{Hella.} & \textbf{PIQA} & \textbf{Wino.} & \textbf{OBQA} & \textbf{Avg.} \\
 & & ppl$_\downarrow$ & ppl$_\downarrow$ & acc$_\uparrow$ & acc$_\uparrow$ & acc$_{\text{n}\uparrow}$ & acc$_{\text{n}\uparrow}$ & acc$_\uparrow$ & acc$_\uparrow$ & acc$_\uparrow$ & acc$_\uparrow$ \\
\shline
\multicolumn{12}{l}{\textit{\textbf{MLA with 10B training tokens and 0.5M batchsize tokens}}} \\
 $\bm{B}\text{Norm}(\bm{A}\bm{x})$
& 342.4M & \textbf{48.73} & \textbf{32.24} & \textbf{30.62} & \textbf{56.99} & \textbf{27.56} & \textbf{36.26} & 63.49 & 51.30 & \textbf{22.00} & \textbf{41.17} \\
 $\bm{B}\bm{A}\bm{x}$
& 342.4M & 61.83 & 32.89 & 28.70 & 54.63 & 26.02 & 35.88 & 63.66 & 51.62 & \textbf{22.00} & 40.36 \\
\rowcolor{Gray}
 $\bm{B}\bm{A}_\text{NoRA-init}\bm{x}$
& 342.4M & 50.77 & 32.42 & 29.87 & 54.04 & 26.37 & 36.11 & \textbf{63.76} & \textbf{52.72} & 21.60 & 40.64 \\
\hline
\multicolumn{12}{l}{\textit{\textbf{MHA with 10B training tokens and 0.5M batchsize tokens}}} \\
 $\bm{W}\bm{x}$
& 348.7M & \textbf{47.95} & \textbf{31.31} & \textbf{30.55} & \textbf{54.59} & \textbf{26.45} & \textbf{36.95} & \textbf{65.07} & \textbf{51.22} & \textbf{22.60} & \textbf{41.06} \\
 $\bm{B}\bm{A}\bm{x}$
& 289.3M & - & - & 0.00 & 25.17 & 28.07 & 25.97 & 49.78 & 49.49 & 16.20 & 27.81 \\
\rowcolor{Gray}
 $\bm{B}\bm{A}_\text{NoRA-init}\bm{x}$
& 289.3M & 63.45 & 34.39 & 28.29 & 54.38 & 26.02 & 35.10 & 63.49 & 50.91 & 19.60 & 39.69 \\
\end{tabular}
\caption{\footnotesize Zero-shot performance of 340M models trained on SlimPajama~\citep{cerebras2023slimpajama}. Commonsense reasoning tasks are evaluated with lm-evaluation-harness~\citep{eval-harness}; the recall-intensive task follows prefix-linear-attention~\citep{arora2024just} with 2K input tokens. ``-'' indicates a collapse in perplexity.}
\label{tab:main_results}
\vspace{-1mm}
\end{table}

\begin{table}[t!]
\centering
\scriptsize
\renewcommand{\arraystretch}{1.22}
\setlength{\tabcolsep}{4.7pt}
\setlength{\abovecaptionskip}{5pt}
\setlength{\belowcaptionskip}{3pt}
\begin{tabular}{lcccccccccc}
\multirow{2}{*}{\textbf{Method}} & \multicolumn{6}{c}{\textbf{Supervised Finetuning}} & \multicolumn{4}{c}{\textbf{Forgetting}} \\
\cmidrule(lr){2-7}\cmidrule(lr){8-11}
 & \textbf{Trainable} & \textbf{GSM8K} & \textbf{Math} & \textbf{HumanEval} & \textbf{MBPP} & \textbf{Avg} & \textbf{MMLU} & \textbf{AGIEval} & \textbf{ARC-C} & \textbf{Avg $\Delta$} \\
\shline
 Base Model
& - & -- & -- & -- & -- & -- & 55.10\pos{0.00} & 23.99\pos{0.00} & 42.92\pos{0.00} & 0.00 \\
 Full
& 3B & \textbf{65.12} & \textbf{17.96} & 36.15 & 49.05 & 42.07 & 54.17\nega{0.93} & \textbf{26.25}\pos{2.26} & 41.64\nega{1.28} & +0.02 \\
 PiSSA
& 48.6M & 54.66 & 12.08 & 39.00 & 55.30 & 40.26 & 54.43\nega{0.67} & 23.99\pos{0.00} & 42.75\nega{0.17} & -0.28 \\
 OFT
& 53.7M & 56.10 & 13.02 & 38.40 & 54.20 & 40.43 & \textbf{54.72}\nega{0.38} & 25.23\pos{1.25} & \textbf{43.60}\pos{0.68} & \textbf{+0.52} \\
 RSLoRA
& 48.6M & 57.05 & 12.32 & 39.65 & \textbf{56.10} & 41.28 & 54.07\nega{1.03} & 24.58\pos{0.60} & 41.81\nega{1.11} & -0.51 \\
 MiSS
& 49.5M & 60.80 & 14.60 & \textbf{39.60} & 55.80 & \textbf{42.70} & 53.40\nega{1.50} & 24.33\pos{0.34} & 41.98\nega{0.94} & -0.70 \\
\hline
\multicolumn{11}{l}{\textit{\textbf{LoRA-style Parameterization}}} \\
 LoRA
& 48.6M & 50.94 & 10.38 & 37.20 & 53.20 & 37.93 & \textbf{54.27}\nega{0.83} & 24.40\pos{0.42} & 41.64\nega{1.28} & -0.56 \\\rowcolor{Gray}
 NoRA-init
& 48.6M & 59.89 & 14.24 & 39.60 & \textbf{55.80} & 42.38 & 54.09\nega{1.01} & 25.10\pos{1.11} & 42.24\nega{0.68} & -0.19 \\\rowcolor{Gray}
 NoRA
& 48.6M & \textbf{61.63} & \textbf{14.46} & \textbf{42.10} & 55.30 & \textbf{43.37} & 54.00\nega{1.10} & \textbf{25.32}\pos{1.33} & \textbf{42.75}\nega{0.17} & \textbf{+0.02} \\
\hline
\multicolumn{11}{l}{\textit{\textbf{DoRA-style Parameterization}}} \\
 DoRA
& 49.4M & 51.63 & 11.36 & 37.80 & 52.40 & 38.30 & \textbf{54.20}\nega{0.90} & 24.09\pos{0.10} & 41.81\nega{1.11} & -0.64 \\\rowcolor{Gray}
 NoRA-init
& 49.4M & 59.59 & 14.10 & 38.70 & 53.20 & 41.40 & 53.65\nega{1.45} & \textbf{24.84}\pos{0.86} & \textbf{42.32}\nega{0.60} & \textbf{-0.40} \\\rowcolor{Gray}
 NoRA
& 49.4M & \textbf{61.33} & \textbf{14.44} & \textbf{42.10} & \textbf{55.60} & \textbf{43.34} & 54.00\nega{1.10} & 24.59\pos{0.58} & 42.15\nega{0.77} & -0.43 \\
\end{tabular}
\caption{\footnotesize Comparison of PEFT methods under supervised finetuning and retained benchmark settings.}
\label{tab:sft_forget_compare}
\vspace{-3.5mm}
\end{table}

\vspace{-0.5mm}
\subsection{Supervised Finetuning}
\vspace{-0.5mm}

Table~\ref{tab:sft_forget_compare} compares NoRA with representative PEFT methods under supervised finetuning. NoRA achieves the best overall SFT performance among the compared methods, improving the average score from 37.93 with standard LoRA to 43.37, a gain of 5.44 points. In particular, NoRA substantially improves performance on GSM8K and HumanEval, reaching 61.63 and 42.10, respectively, while also outperforming PiSSA~\citep{meng2024pissa}, OFT~\citep{qiu2023oft,qiu2025oftv2,liu2024boft}, RSLoRA~\citep{kalajdzievski2023rank}, and MiSS~\citep{kang2026miss} in average performance. These results demonstrate the effectiveness of rank-dimension normalization for improving LoRA.

\begin{figure}[t]
    \centering
    \vspace{-1mm}
    \includegraphics[width=\linewidth]{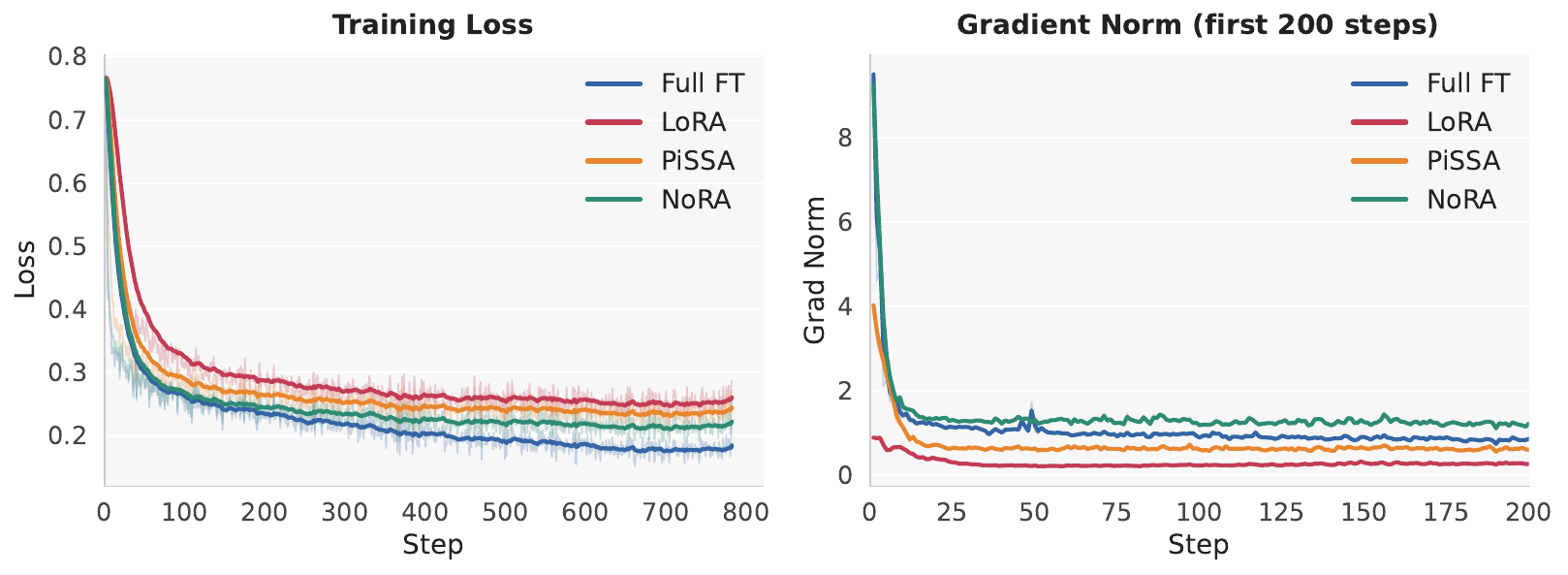}
    \vspace{-6.5mm}
    \caption{\footnotesize Training dynamics comparison of full finetuning, LoRA, PiSSA, and NoRA on the SFT task.}
    \label{fig:trajectory}
    \vspace{-2mm}
\end{figure}

We further compare NoRA with its initialization-only variant, NoRA-init. NoRA-init applies normalization constraint only to the initial down-projection and then follows standard LoRA training, whereas NoRA continuously enforces the normalization constraint throughout training. NoRA-init already improves the SFT average from 37.93 to 42.38, capturing a large portion of the gain achieved by NoRA. With the normalization constraint maintained throughout training, NoRA further improves the average score to 43.37. The comparison indicates that controlling the input-to-latent projection magnitude at initialization accounts for a large portion of the gain, while persistent normalization provides additional improvements. Comparison of training dynamics is given in Figure~\ref{fig:trajectory}. NoRA shows stronger learnability than LoRA without introducing extra parameters.

NoRA also outperforms MiSS, which achieves an average score of 42.70. Recall that MiSS can be reformulated as a fixed block-identity down-projection satisfying the normalization constraint. The strong performance of both MiSS and NoRA-init therefore provides further evidence that normalized low-rank projections play an important role in their optimization. Notably, NoRA-init achieves comparable performance to MiSS while keeping the down-projection fully trainable, avoiding the constraint of fixing the projection throughout training.

Finally, we examine whether the NoRA principle generalizes beyond the standard LoRA parameterization. We apply $\text{Norm}_r$-based normalization to DoRA and obtain consistent improvements: NoRA-init-DoRA improves the SFT average from 38.30 for DoRA to 41.40. This result demonstrates that the benefit of rank-dimension normalization is not specific to LoRA, but can also be transferred to other low-rank adaptation variants. Together with the improvements observed for LoRA, this suggests that NoRA captures a general optimization principle for low-rank adaptation. NoRA also exhibits favorable knowledge retention, achieving an average change of +0.02 on MMLU, AGIEval, and ARC-C, compared with $-0.56$ for LoRA and -0.70 for MiSS. Thus, NoRA improves adaptation performance while maintaining the pretrained model's retained knowledge.

\definecolor{maingain}{RGB}{0,102,51}
\definecolor{mainloss}{RGB}{160,32,48}
\newcommand{\ppdelta}[1]{%
  \begingroup
  \edef\ppdeltatmp{#1}%
  \ifnum\pdfstrcmp{\ppdeltatmp}{+0.00}=0
    \textcolor{black!65}{(#1)}%
  \else
    \expandafter\ppdeltaaux\ppdeltatmp\relax
  \fi
  \endgroup
}
\def\ppdeltaaux#1#2\relax{%
  \if#1+%
    \textcolor{maingain}{(#1#2)}%
  \else
    \textcolor{mainloss}{(#1#2)}%
  \fi
}

\vspace{-0.5mm}
\subsection{RLVR on Mathematical Reasoning}
\vspace{-0.5mm}

We further evaluate NoRA under reinforcement learning with verifiable rewards (RLVR) on challenging mathematical reasoning benchmarks. As shown in Table~\ref{tab:rl_results}, standard LoRA improves the overall average score from 41.0 for the base model to 42.8, while NoRA further increases it to 44.4, outperforming the base model and standard LoRA by 3.4 and 1.6 points, respectively. NoRA yields broad improvements across the evaluated benchmarks, with particularly clear gains on AMC, MATH500, and Minerva. These results suggest that rank-dimension normalization remains effective under RLVR and facilitates more effective optimization of low-rank adapters.

A notable difference emerges for initialization methods based on pretrained-weight spectral decomposition. Although PiSSA~\citep{meng2024pissa} and MiLoRA~\citep{wang2025milora} are effective under supervised finetuning, their performance degrades substantially in the RLVR setting. This behavior is consistent with prior observations that spectral initialization methods can suffer from optimization instability during reinforcement learning~\citep{yin2025evaluatingparameterefficientmethods}. In contrast, NoRA remains stable and consistently improves the overall performance of standard LoRA without relying on singular-value decomposition of the pretrained weights.
These results highlight the robustness of NoRA across different optimization regimes. Unlike PiSSA and MiLoRA, NoRA does not depend on the spectral structure of pretrained weights, while preserving the lightweight low-rank parameterization and deployment efficiency of standard LoRA. Together with the supervised finetuning results, the RLVR experiments suggest that controlling the scale of the input-to-latent projection provides a simple and broadly effective mechanism for improving low-rank optimization.

\begin{table}[t]
\centering
\scriptsize
\renewcommand{\arraystretch}{1.22}
\setlength{\tabcolsep}{7pt}
\setlength{\abovecaptionskip}{5pt}
\setlength{\belowcaptionskip}{3pt}
\begin{tabular}{lccccccccccccc}
\multirow{2}{*}{\textbf{Method}} & \multicolumn{2}{c}{\textbf{AIME24@32}} & \multicolumn{2}{c}{\textbf{AIME25@32}} & \multicolumn{2}{c}{\textbf{AMC@32}} & \multicolumn{2}{c}{\textbf{HMMT@32}} & \multicolumn{2}{c}{\textbf{MATH500@4}} & \multicolumn{2}{c}{\textbf{Minerva@4}} & \multirow{2}{*}{\textbf{Avg.}}\\
 & Avg. & Pass & Avg. & Pass & Avg. & Pass & Avg. & Pass & Avg. & Pass & Avg. & Pass &\\
\shline
 Base
 & 24.3 & 70.0 & 20.4 & \textbf{53.3} & 64.8 & 95.0 & 9.9 & \textbf{36.7} & 80.9 & \textbf{92.8} & 27.8 & 43.0 & 41.0\\
 MiLoRA
 & 4.2 & 6.7 & 0.0 & 0.0 & 19.6 & 47.5 & 0.0 & 0.0 & 44.5 & 63.4 & 11.7 & 19.9 & 18.0\\
 PiSSA
 & 0.0 & 0.0 & 0.0 & 0.0 & 0.0 & 0.0 & 0.0 & 0.0 & 0.6 & 1.0 & 0.1 & 0.4 & 0.2\\
 LoRA
 & \textbf{28.0} & 60.0 & 23.1 & \textbf{53.3} & 68.0 & \textbf{97.5} & \textbf{10.3} & 35.0 & 81.3 & 92.2 & 30.1 & 46.3 & 42.8\\
\rowcolor{Gray}
 NoRA
 & 26.7 & \textbf{66.7} & \textbf{23.5} & 46.7 & \textbf{72.7} & \textbf{97.5} & \textbf{10.3} & 35.0 & \textbf{84.5} & \textbf{92.8} & \textbf{32.0} & \textbf{47.8} & \textbf{44.4}\\
\end{tabular}
\caption{\footnotesize Comparison of accuracy and pass rate on the RLVR task. All numbers are reported in percentages.}
\label{tab:rl_results}
\vspace{-3mm}
\end{table}

\vspace{-0.5mm}
\section{Related Work and Concluding Remarks}
\vspace{-0.5mm}

\textbf{Parameter-efficient finetuning.}
As LLMs continue to grow in scale and capability, there has been increasing interest in adapting them to downstream tasks in a parameter-efficient manner~\citep{houlsby2019parameter,aghajanyan2020intrinsic,hu2021lora,edalati2022krona,wang2022adamix,gheini2021cross,zaken2022bitfit,guo2020parameter,sung2021training,ansell2022composable,lester2021power,li2021prefix,vu2022spot,he2021towards,mao2021unipelt,karimi2021compacter,liu2022few,sung2022lst,chen2022parameter,jia2022visual,chen2022adaptformer,zhang2022neural,jie2023fact,lian2022scaling,qiu2023oft,qiu2025oftv2,liu2024boft,wu2024mixture,zi2023delta}. 
Among them, LoRA~\citep{hu2021lora} is widely adopted due to its simplicity, efficiency, and mergeability at inference time. 
Subsequent variants improve LoRA from different perspectives, such as adaptive rank allocation~\citep{zhang2023adalora}, low-bit quantization~\citep{dettmers2023qlora} and weight-decomposed low-rank updates (DoRA)~\citep{liu2024dora}.
While these methods improve the capacity, efficiency, or stability of low-rank adaptation, our work focuses on a complementary question: how the low-rank subspace should be initialized for better optimization.

\textbf{LoRA initialization and optimization.}
Recent studies show that LoRA is sensitive to initialization, scaling, and early-stage optimization dynamics. PiSSA initializes LoRA from principal singular components of pretrained weights~\citep{meng2024pissa}, OLoRA~\citep{buyukakyuz2024olora} adopts orthogonal initialization, LoRA-GA~\citep{wang2024lora} uses gradient information to guide initialization, EVA leverages activation statistics, and rsLoRA improves stability through rank-dependent scaling~\citep{kalajdzievski2023rank}. These methods demonstrate the importance of initialization, but often rely on SVD, gradient estimation, activation statistics, or scaling heuristics. In contrast, NoRA requires no additional data, SVD, or test-time overhead while preserving the mergeability of LoRA.

\textbf{Concluding remarks.} This work shows that LoRA’s effectiveness depends not only on its rank, but also critically on the geometry and scale of its down-projection. Inspired by the normalized latent representations in MLA, we ask whether the benefits of latent normalization can be transferred to LoRA without sacrificing its linear structure and exact mergeability. This leads to NoRA, which normalizes the down-projection along the rank dimension and thereby controls the scale of the input-to-latent mapping. From a preconditioning perspective, LoRA can be viewed as full finetuning under an implicit low-rank, input-side preconditioner determined by the down-projection; NoRA's rank-dimension normalization corrects undesirable scale imbalance in this preconditioner and yields a better-conditioned optimization geometry. NoRA can consistently improve convergence, training stability and downstream performance across pretraining, supervised finetuning, and RLVR.

\bibliography{iclr2026_conference}
\bibliographystyle{iclr2026_conference}
\appendix
\newpage
\section{Pretraining Settings}

We conduct the pre-training experiments on FineWeb-10BT. Unless otherwise
specified, all models are trained for 20,480 optimization steps with a
sequence length of 2,048 and a global batch size of 256. We use AdamW with
a peak learning rate of $3\times10^{-4}$ and $\epsilon=10^{-15}$. The
learning rate is warmed up for the first 1,024 steps and then decayed using
a cosine schedule to $10\%$ of the peak learning rate. The gradient norm is
clipped at 1.0. All experiments use the same training configuration and
random seed for controlled comparison.

\begin{table}[h]
\centering
\scriptsize
\renewcommand{\arraystretch}{1.26}
\setlength{\tabcolsep}{8pt}
\setlength{\abovecaptionskip}{5pt}
\setlength{\belowcaptionskip}{3pt}
\begin{tabular}{lc}
\textbf{Hyperparameter} & \textbf{Value} \\
\shline
Dataset & FineWeb-10BT \\
Optimizer & AdamW \\
Learning rate & $3\times10^{-4}$ \\
Adam $\epsilon$ & $10^{-15}$ \\
LR scheduler & Cosine decay \\
Warmup steps & 1,024 \\
Minimum LR ratio & 0.1 \\
Global batch size & 256 \\
Sequence length & 2,048 \\
Training steps & 20,480 \\
Gradient clipping & 1.0 \\
Random seed & 42 \\
\end{tabular}
\caption{\footnotesize Hyperparameter settings for the pretraining experiments.}
\label{tab:pretrain_settings}
\vspace{-1mm}
\end{table}

\section{Supervised Finetuning Settings}\label{app:settings}
For the SFT experiments, we tune the hyperparameters of each method individually to ensure competitive performance. For NoRA, we recommend using $\alpha=r$ as the default setting, under which the early gradient norm is close to that of full finetuning. Although a larger scaling factor, such as $\alpha=2r$, can yield better loss fitting in some cases, we observe that it may also lead to increased forgetting of the pretrained knowledge. Considering this trade-off between adaptation and retention, we use and recommend $\alpha:r=1:1$ as the default configuration.

\begin{table}[h]
\centering
\scriptsize
\renewcommand{\arraystretch}{1.26}
\setlength{\tabcolsep}{8pt}
\setlength{\abovecaptionskip}{5pt}
\setlength{\belowcaptionskip}{3pt}
\begin{tabular}{lcccccc}
\textbf{Hyperparameters} & \textbf{LoRA} & \textbf{DoRA} & \textbf{PiSSA} & \textbf{rsLoRA} & \textbf{OFT} & \textbf{NoRA} \\
\shline
rank & 32 & 32 & 32 & 32 & 32 & 32 \\
$\alpha$ & 64 & 64 & 32 & 64 & - & 32 \\
\hline
dropout & \multicolumn{6}{c}{0.0} \\
optimizer & \multicolumn{6}{c}{AdamW} \\
lr & \multicolumn{6}{c}{2e-5} \\
lr scheduler & \multicolumn{6}{c}{Cosine decay} \\
batch size & \multicolumn{6}{c}{128} \\
warmup ratio & \multicolumn{6}{c}{0.3} \\
epochs & \multicolumn{6}{c}{1} \\
target\_modules & \multicolumn{6}{c}{Q,K,V,O,Up,Down,Gate} \\
\end{tabular}
\caption{\footnotesize Hyperparameter settings for the Llama-3.2-3B supervised finetuning experiments.}
\label{tab:llava_hyperparamters}
\vspace{-1mm}
\end{table}

\newpage
\section{Reinforcement Learning Settings}

We conduct reinforcement learning experiments on
DeepSeek-R1-Distill-Qwen-1.5B using the DAPO objective and the
DAPO-Math-17K dataset. All experiments are performed with 8 GPUs using
bfloat16 precision. We train for 1,024 optimization steps with a global
batch size of 128 and a learning rate of $1\times10^{-5}$ under a cosine
learning-rate schedule without warmup. For each prompt, we sample 8
responses with a maximum completion length of 16,384 tokens. The maximum
prompt length is set to 512 tokens.

For parameter-efficient training, we use a rank of 32 and set
$\alpha=64$. The low-rank modules are applied to all attention projections
($q$, $k$, $v$, and $o$) and MLP projections (up, down, and gate). We use
the same training configuration across different methods to ensure a
controlled comparison.

\begin{table}[h]
\centering
\scriptsize
\renewcommand{\arraystretch}{1.26}
\setlength{\tabcolsep}{8pt}
\setlength{\abovecaptionskip}{5pt}
\setlength{\belowcaptionskip}{3pt}
\begin{tabular}{lc}
\textbf{Hyperparameter} & \textbf{Value} \\
\shline
Model & DeepSeek-R1-Distill-Qwen-1.5B \\
Dataset & DAPO-Math-17K \\
RL objective & DAPO \\
Precision & bfloat16 \\
Learning rate & $1\times10^{-5}$ \\
LR scheduler & Cosine decay \\
Warmup ratio & 0 \\
Global batch size & 128 \\
Training steps & 1,024 \\
LoRA rank $r$ & 32 \\
LoRA $\alpha$ & 64 \\
LoRA dropout & 0.05 \\
Responses per prompt & 8 \\
Maximum prompt length & 512 \\
Maximum completion length & 16,384 \\
$\epsilon_{\mathrm{high}}$ & 0.28 \\
$\beta$ & 0.0 \\
Random seed & 42 \\
\end{tabular}
\caption{\footnotesize Hyperparameter settings for the reinforcement learning experiments.}
\label{tab:rl_settings}
\vspace{-1mm}
\end{table}
\end{document}